\pdfoutput=1

\documentclass[11pt]{article}

\usepackage{acl}

\usepackage{times}
\usepackage{latexsym}
\usepackage{graphicx}

\usepackage[T1]{fontenc}

\usepackage[utf8]{inputenc}

\usepackage{microtype}

\newcommand{\corp}{{\sc InferES}~}

\title{\corp : A Natural Language Inference Corpus for Spanish \\ Featuring Negation-Based Contrastive and Adversarial Examples}

\author{Venelin Kovatchev \\
  School of Information \\
  The University of Texas at Austin \\
  \texttt{venelin@utexas.edu} \\\And
  Mariona Taul\'{e} \\
  Centre de Llenguatge i Computaci\'{o} \\
  Institut de Recerca en Sistemes Complexos \\
  Universitat de Barcelona \\
  \texttt{mtaule@ub.edu} \\}

\begin{document}
\maketitle
\begin{abstract}
In this paper, we present \corp - an original corpus for Natural Language Inference (NLI) in European Spanish. We propose, implement, and analyze a variety of corpus-creating strategies utilizing expert linguists and crowd workers. The objectives behind \corp are to provide high-quality data, and, at the same time to facilitate the systematic evaluation of automated systems. Specifically, we focus on measuring and improving the performance of machine learning systems on negation-based adversarial examples and their ability to generalize across out-of-distribution topics.

We train two transformer models on \corp (8,055 gold examples) in a variety of scenarios. Our best model obtains 72.8\% accuracy, leaving a lot of room for improvement. The ``hypothesis-only'' baseline performs only 2\%-5\% higher than majority, indicating much fewer annotation artifacts than prior work. We find that models trained on \corp generalize very well across topics (both in- and out-of-distribution) and perform moderately well on negation-based adversarial examples.
\end{abstract}

\section{Introduction}\label{inferes:intro}

In the task of Natural Language Inference (NLI), an automated system has to determine the meaning relation that holds between two texts. The model has to make a three-way choice between \textit{entailment}: a hypothesis (h) is \underline{true} given a premise (p) (e.g. \textbf{1.}); \textit{contradiction}: a hypothesis (h) is \underline{false} given a premise (p) (e.g. \textbf{2.}); or \textit{neutral}: the truth value of the hypothesis (h) cannot be determined solely based on the premise (p) (e.g.: \textbf{3.}).

\begin{itemize}

    \item[\textbf{1.}] \textit{p)}~John goes to work every day with a car. \\
    \textit{h)}~John has a job.
    \item[\textbf{2.}] \textit{p)}~John goes to work every day with a car.\\
    \textit{h)}~John takes the bus to go to work.
    \item[\textbf{3.}] \textit{p)}~John goes to work every day with a car. \\
    \textit{h)}~John has a Porsche.

\end{itemize}

NLI (formerly known as Recognizing Textual Entailment (RTE)) is one of the core tasks in the popular benchmarks for Natural Language Understanding GLUE \cite{wang-etal-2018-glue} and Super GLUE \cite{NEURIPS2019_4496bf24}. Hundreds of machine learning systems compete on these benchmarks, improving the state of NLU.

One key limitation of NLI research is that most of the existing corpora are only for English. Limited research has been done on multilingual and non-English corpora \cite{10.1007/11671299_29,conneau-etal-2018-xnli,https://doi.org/10.48550/arxiv.2009.08820,ham-etal-2020-kornli,hu-etal-2020-ocnli,mahendra-etal-2021-indonli}.


Another well-known issue with NLI is the quality of the existing datasets and the limitations of the models trained on them.
On most NLI corpora, state-of-the-art transformer based models can obtain quantitative results (Accuracy and F1) that equal or exceed human performance. Despite this high performance, 
researchers have identified numerous limitations and potential problems. \citet {poliak-etal-2018-hypothesis} found that annotation artifacts in the datasets enable the models to predict the label by only looking at the hypothesis. NLI models are often prone to adversarial attacks \cite{williams-etal-2018-broad} and may fail on instances that require specific linguistic capabilities \cite{hossain-etal-2020-analysis,saha-etal-2020-conjnli}.

In this paper we address both of these shortcomings in NLI research. We present \corp - to the best of our knowledge, the first original NLI corpus for Spanish, not adapted from another language or task. We study prior work for strategies that can reduce annotation artifacts and increase the linguistic variety of the corpus, resulting in a dataset that is more challenging for automated systems to solve. We also design the corpus in a way that facilitates systematic evaluation of automated systems on: 1) negation-based adversarial examples; 2) out-of-distribution examples.


We propose, implement, and analyze three different strategies for the generation and annotation of text pairs. In the \textit{generation} strategy, expert linguists write original hypotheses given a premise. In the \textit{rewrite} strategy, expert linguists create contrastive and adversarial examples by rewriting and re-annotating ``generated'' pairs. 
In the \textit{annotation} strategy, we first generate text pairs in a semi-automated manner and then use crowd annotators to determine the meaning relation. The final \corp corpus contains 8,055 gold standard premise-hypothesis pairs. The core part of the corpus is expert-generated and we make an additional effort to ensure the quality of the data and the linguistic diversity of the examples.

We provide two baseline for \corp by finetuning multilingual BERT and BETO (Spanish BERT) transformer models. On the full dataset, BETO obtains 72.8\% accuracy, indicating that the classification task is non-trivial. Both mBERT and BETO perform poorly in the ``hypothesis-only'' condition, indicating fewer annotation artifacts in the corpus compared to prior work. Both systems generalize well across the different topics in \corp both ``in-distribution'' and ``out-of-distribution''. We notice a substantial drop in performance when evaluating negation-based adversarial examples, however the systems still outperform majority and ``hypothesis-only''.

\corp expands the scope of the NLI research in Spanish, provides new set of naturally occurring contrastive and adversarial examples, and facilitates the study of negation and coreference in the context of NLI. As part of the corpus creation, we also present and analyze three unique strategies for creating examples. All our data and baseline models are being  released to the community\footnote{At \url{https://github.com/venelink/inferes} InferES is also added as a HuggingFace dataset}.

The rest of this article is organized as follows. Section \ref{inferes:rl} discusses the related work. Section \ref{inferes:corp} formulates our objectives and introduces the different corpus-creation strategies. Section \ref{inferes:cs} describes the final corpus and basic statistical data regarding it. Section \ref{inferes:exp} presents the machine learning experimental setup and results. Section \ref{inferes:diss} is devoted to a discussion of the results and their implications. Finally, Section \ref{inferes:conc} concludes the article. 

\section{Related Work}\label{inferes:rl}

The task of Recognizing Textual Entailment (RTE) was proposed in \citet{dagan2006pascal} as a binary classification (``entailment'' / ``non-entailment''). The RTE competition ran for seven editions \cite{bar2006second,giampiccolo2007third,giampiccolo2008fourth,bentivogli2009fifth,Bentivogli2010TheSP,bentivogli2011seventh}. RTE was later reformulated as a three-way decision and ultimately renamed Natural Language Inference in the SNLI \cite{bowman2015large} and the MNLI \cite{williams-etal-2018-broad} corpora. Both the RTE and the NLI tasks form part of the Natural Language Understanding benchmarks GLUE \cite{wang-etal-2018-glue} and Super-GLUE \cite{NEURIPS2019_4496bf24}. The NLU benchmarks attracted a lot of attention from the community and by 2020 the state-of-the-art systems reported human level performance. \citet{parrish-etal-2021-putting-linguist} proposed a ``linguist-in-the-loop'' corpus creation to improve the quality of the data.

The ``super-human'' performance of NLI systems has been questioned by a number of researchers.
\citet{poliak-etal-2018-hypothesis} found that annotation artifacts in the datasets enable the models to predict the label by only looking at the hypothesis. \citet{mccoy-etal-2019-right} and \citet{gururangan-etal-2018-annotation} demonstrate that 
state-of-the-art NLI systems often rely on heuristics and annotation artifacts.

Systematic approaches to evaluation propose different sets of stress-tests for NLI and NLU systems \cite{kovatchev-etal-2018-etpc,naik-etal-2018-stress,wallace-etal-2019-universal,kovatchev-etal-2019-qualitative,ribeiro-etal-2020-beyond,kovatchev-etal-2020-decomposing}. The attacks can be inspired by linguistic phenomena or empirical use cases.
Systematic evaluations show that NLI and other NLU systems often underperform on complex linguistic phenomena such as conjunction \cite{saha-etal-2020-conjnli}, negation \cite{hossain-etal-2020-analysis}, and coreference \cite{kovatchev-etal-2022-longhorns}.
Researchers also experimented with creating contrastive examples that differ only slightly from training examples, but have a different label \cite{glockner-etal-2018-breaking,Kaushik2020Learning,gardner-etal-2020-evaluating}. Adversarially created datasets such as Adversarial NLI \cite{nie-etal-2020-adversarial} and Dynabench NLI \cite{kiela-etal-2021-dynabench} demonstrate that there is a lot of room for improvement regarding NLI datasets and models.

Most of the available resources for NLI research are in English. \citet{conneau-etal-2018-xnli} present XNLI, a multilingual dataset created by translating English NLI examples into other languages. The interest in multilingual NLI has resulted in the creation of some novel non-English resources such as the Korean NLI corpus \cite{ham-etal-2020-kornli}, Chinese NLI corpus \cite{hu-etal-2020-ocnli}, Persian NLI corpus \cite{https://doi.org/10.48550/arxiv.2009.08820}, Indonesian NLI corpus \cite{mahendra-etal-2021-indonli}, and indigenous languages of the Americas NLI corpus \cite{ebrahimi-etal-2022-americasnli}. For Spanish, the only available resources are the Spanish portion of XNLI and the SPARTE corpus for RTE \cite{10.1007/11671299_29} which was adapted from Question Answering data.

\section{Objectives and Corpus Creation}\label{inferes:corp}

When creating \corp we experimented with different strategies for obtaining gold examples. To the best of our knowledge, this is the first time various annotation strategies are combined and compared in a single NLI corpus. We adopt three different approaches, used in prior work: our \textbf{generation} strategy is similar to the original RTE and NLI corpus creation; our \textbf{rewrite} strategy is inspired from work in generating adversarial and contrastive examples; our \textbf{annotation} strategy scales well with data and allows us to compare expert- and crowd-crated datasets. Our aim was to provide interesting and diverse examples that cover a large range of use cases and linguistic phenomena. We hope that \corp can be used not only to train automated systems, but also to better understand the nature of inference. We formulated three main \textbf{objectives}:

\begin{itemize}
    \item[O1] To create a native NLI dataset for the Spanish language. The existing resources are either an adaptation from a different task or a translation from English.
    \item[O2] To promote better data quality and corpus creation practices. We aim to create a more challenging dataset and simultaneously reduce the number of annotation artifacts.
    \item[O3] To facilitate the research on negation and coreference in the context of NLI. More specifically, we focus on contrastive and adversarial examples.
\end{itemize}


\subsection{Premise Extraction}\label{i:corp:ext}

In the first step of the process, we extracted a set of candidate premises. We decided to use a single sentence premise, similar to SNLI and MNLI datasets. We defined two requirements for our premise sentences: 1) that they cover a range of different topics; and 2) that they be complex enough to entail or contradict multiple possible hypotheses.

\paragraph{Choice of topics} As a source for premises we used the Spanish version of Wikipedia from October 2019. We chose six topics, covering five different domains: history, art, sports, technology, and politics. We also selected the topics in pairs hypothesizing that this selection might facilitate the creation of contrastive examples, specifically in the context of coreference.
\begin{itemize}
    \item famous historical figures: \\ 
    Pablo Picasso (ES: Pablo Picasso) \\  
    Christopher Columbus (ES: Cristobal Col\'{o}n)

    \item types of ``games'': \\
    Olympic games (ES: Juegos Ol\'{i}mpicos) \\ 
    Video games (ES: videojuegos)
    \item types of multinational ``unions'': \\
    The European Union (ES: Uni\'{o}n Europeo) \\ 
    The Union of Soviet Socialist Republics \\ (ES: Uni\'{o}n Sov\'{e}tica)
\end{itemize}

\paragraph{Extraction process} 
We extracted the main Wikipedia article for each topic and preprocessed it (sentence segmentation and tokenization) using Spacy \cite{spacy2}. We split the text by paragraphs and discarded paragraphs that contained only one sentence or more than five sentences. Then, from each paragraph, we selected a single sentence, prioritizing sentences containing negation\footnote{To check for negation, we used a simple keyword based search, using a list of the most common negative particles, adverbs, and verbs in Spanish. The list is available at \url{https://github.com/venelink/inferes}} where possible, otherwise selecting a sentence at random. We ensured that each selected sentence had a length between 15 and 45 tokens. 

\paragraph{Post-processing}
At the end of the extraction process, we had 471 candidate-premise sentences as follows: 82 for Picasso, 60 for Columbus, 68 for the Olympic games, 73 for video games, 107 for the EU, and 81 for the USSR. For each sentence, we also kept the corresponding paragraph to enable experimental setup where we provide an additional context to the machine learning models at train and test time. We also used the ``context paragraphs'' when generating ``neutral'' pairs. One of the authors manually inspected all 471 candidate-premise sentences. They manually resolved problems with sentence segmentation, removed URLs and internal wikipedia document references, and explicitly resolved any coreferential and anaphorical ambiguities (i.e.: replaced pronouns and coreferential entities with an unambiguous referent). 

\subsection{Expert ``Generation'' Strategy}\label{i:corp:gen}

\paragraph{Task formulation}
We formulated two separate \textit{generation} tasks: the generation of entailment pairs and generation of contradiction pairs. We defined the tasks as follows:

\textbf{Entailment:} \textit{Given a premise, write two different sentences that are true.}

\textbf{Contradiction:} \textit{Given a premise, write two different sentences that are false.} \\
Our guidelines enforced a strict definition of contradiction and required our generators to write sentences that explicitly contradict the premise, rather than implicitly rely on event and actor coreference\footnote{For a discussion of the definition of contradiction in the context of NLI, we refer the reader to \citet{gold-etal-2019-annotating}.}. We asked the generators to provide multiple examples, requiring the use of different strategies. We further instructed the corpus generators to: 1) generate hypotheses that have a low lexical overlap with the premise; 2) generate one affirmative and one negated sentence for each relation; 3) where possible, replace named entities with pronouns or other instances of coreference. Our instructions aim to encourage generators to come up with difficult and diverse examples. 
We also ensure a high frequency of entailment pairs containing negation and of affirmative contradiction pairs. For a reference, the readers can see an example of generated entailment and contradiction hypotheses in \textbf{4.} 

\begin{itemize}
    \item[\textbf{4.}] \textsc{(premise)} En la d\'{e}cada de 1980 el soporte habitual para el software era el cartucho en las videoconsolas, y el disco magn\'{e}tico o la cinta de casete en los ordenadores.\footnote{EN: ``In the 1980s, the usual medium for software was the cartridge in video consoles, and the magnetic disk or cassette tape in computers.''}
    
    \textsc{(entailment)} Es poco probable que en los 1980s las videoconsolas y los ordenadores utilizaran el mismo soporte.\footnote{EN: ``It is unlikely that in the 1980s video game consoles and computers used the same medium.''}

    \textsc{(contracition)} Aunque el cartucho se hab\'{i}a utilizado en el pasado, en los 80 ya se consideraba desfasado.\footnote{EN: ``Although the cartridge had been used in the past, by the 1980s it was already considered outdated.''}
\end{itemize}


\paragraph{Sentence ``generators''}
Four graduate students of linguistics were trained for this task by the authors of the paper. They received detailed instructions, examples, and a two-hour interactive training session prior to the start of the corpus creation. The students met with the authors of the paper on a weekly basis to discuss challenging or interesting examples. To further increase the diversity of the corpus, we recruited 24 undergraduate students for a  two-hour hypothesis generation session preceded by a one-hour interactive training session. All generators were native speakers of European Spanish.

\paragraph{``Generated'' portion of the corpus}
We distributed the premises extracted in Section \ref{i:corp:ext} between the four graduate students balancing the number of entailment and contradiction pairs per topic and per premise. For any given premise, a single expert would generate only one of the relations, never both. Some premises were used more than once. Our selection strategy aimed to create maximum diversity in the data and reduce the potential bias from a relatively small number of data generators. The four graduate students created 2,284 pairs from the original 471 premises. The 24 undergraduate students generated further 872 pairs. The final corpus from the \textit{generation} strategy contains 3,156 pairs, split equally between entailment and contradiction. We describe the process of obtaining ``neutral'' text pairs in Section \ref{i:c:neu}.




\begin{figure*}[t!]
    \centering
    \includegraphics[width=0.95\textwidth]{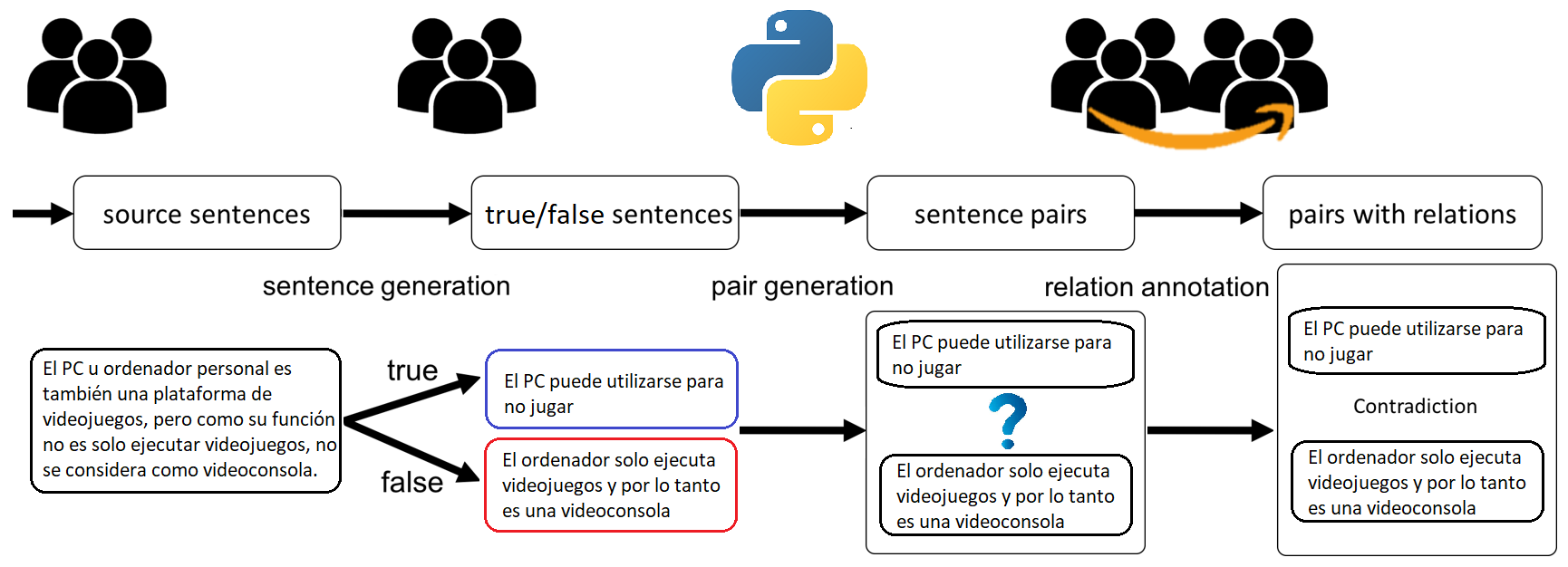}
    \caption{Pipeline for the \textit{annotation} corpus creation strategy.}
    \label{i:fig:crowd}
\end{figure*}

\subsection{Expert ``Rewrite'' Strategy}\label{i:corp:rew}

\paragraph{Task formulation} The \textit{rewrite} strategy is based on the pairs from the \textit{generation} strategy. We defined the task as follows:

\textit{Given an existing premise--hypothesis pair, modify both the premise and the hypothesis so that:}

\textit{1. the resulting sentence has a substantial difference in meaning from original}

\textit{2. where possible, change the negation status of a sentence. That is, an affirmative sentence would become negated, while a negated sentence would become affirmative}

\textit{3. where possible, replace some words in the original sentences with synonyms and/or coreferential entities}

We further instructed the ``rewriters'' not to resort only to simple negation. An example of the \textit{rewrite} strategy can be seen in \textbf{5.} and \textbf{6.}: when rewriting the premise, our expert replaced ``descartaba'' (EN:``ruled out'') with ``aceptaba'' (EN: ``accepted''); when rewriting the hypothesis, they changed ``inaceptable'' (EN: ``unacceptable'') for ``viable'' (EN: ``feasible''). The resulting adversarial examples include lexical and morphological negation and are more complex than the ``simple negation'' benchmark of \citet{hossain-etal-2020-analysis}.

\begin{itemize}
    \item[\textbf{5.}] \textsc{(Pr)} La reina llam\'{o} entonces a Col\'{o}n, dici\'{e}ndole que no \textbf{descartaba} totalmente su plan.\footnote{EN: The queen then called Columbus, telling him that she did not fully \textbf{rule out} his plan.}

    \textsc{(Hyp)} La reina le hizo saber a Col\'{o}n que su plan no era del todo \textbf{inaceptable}.\footnote{EN: The queen let Columbus know that his plan was not entirely \textbf{unacceptable}.}

    \item[\textbf{6.}] \textsc{(Pr\_rw)} La reina llam\'{o} entonces a Col\'{o}n, dici\'{e}ndole que no \textbf{aceptaba} totalmente su plan.\footnote{EN: The queen then called Columbus, telling him that she did not fully \textbf{accept} his plan.}
    
    \textsc{(Hyp\_rw)} La reina le hizo saber a Col\'{o}n que su plan no era del todo \textbf{viable}.\footnote{EN: The queen let Columbus know that his plan was not entirely \textbf{feasible}.}
\end{itemize}

\paragraph{``Rewritten'' portion of the corpus} The \textit{rewrite} process was carried out by two graduate students of linguistics. After rewriting both the premise and the hypothesis, we create three new combinations involving an original or rewritten hypothesis. In the provided example, those are the pairs 5.(Pr)--6.(Hyp\_rw), 6.(Pr\_rw)--5.(Hyp) and 5.(Hyp)--6.(Hyp\_rw).\footnote{We do not to use 5.(Pr)--6(Pr\_rw) due to sentence length.} Our ``rewriters'' then annotated the relations between the new pairs (in the example, the relations are ``contradiction'', ``neutral'', and ``entailment'' respectively). As a souorce, we selected 20 entailment and 20 contradiction per topic, a total of 240 ``generated'' pairs. We ensured equal distribution of the original ``generators'' and created 720 new adversarial ``rewrite'' pairs.

\subsection{Crowd ``Annotation'' Strategy}\label{i:corp:cs}

\paragraph{Task formulation} 
For the crowd \textit{annotation} strategy we adopted the three-step approach proposed by \citet{gold-etal-2019-annotating}. The authors first semi-automatically generated a large pool of premise-hypothesis pairs. Then, they applied stratified sub-sampling. Finally, they recruited crowd workers to annotate the meaning relations between the texts. Figure \ref{i:fig:crowd} illustrates the \textit{annotation} strategy. We choose this approach since it's compatible with our \textit{generation} and \textit{rewrite} strategies. We were interested in comparing and combining expert- and crowd-created corpora, which, to the best of our knowledge has not been done before for NLI.

\paragraph{Creating a sentence pool}
The first step of the process was identical to the \textit{generation} strategy. We chose 20 of the original premises, five from Picasso, Columbus, Olympic games, and video games. We chose premises that contain multiple predicates and would allow for creativity in generating entailment and contradiction pairs. In Figure \ref{i:fig:crowd}, these premises are called ``source sentences''. \footnote{EN: ``The PC or personal computer is also a video game platform, but since its function is not only to play video games, it is not considered a video game console.''} 

We recruited 26 undergraduate students of linguistics and provided them with one-hour interactive training for the task of generating entailment and contradiction pairs. Each student generated 20 entailment and 20 contradiction pairs from the same 20 ``source sentences''. In Figure \ref{i:fig:crowd}, these hypotheses are called ``true/false sentences''. \footnote{EN: (true) ``The PC can be used for non-gaming.''}\footnote{EN: (false) ``The computer only runs video games and is therefore a game console.'' } At this step, the students generated a ``sentence pool'' of of 1,040 ``true/false sentences''.

\paragraph{Pair generation}In the second step we combined the sentences from the ``sentence pool'' in pairs using three different selection strategies. The ``true-true'' strategy combines two ``true'' sentences derived from the same premise. The ``false-false'' strategy combines two ``false'' sentences from the same premise. The ``true-false'' strategy combines one true and one false sentence derived from the same premise. Unlike \citet{gold-etal-2019-annotating}, we do not include a random pairing and do not downsample ``false-false'' and ``true-false'' strategies. We randomly selected 2,000 of the pairs for annotation, ensuring equal distribution of strategies.

\paragraph{Pair annotation}
In the third step, we asked crowd workers to annotate the textual relation between pairs. We used the WARP-Text \cite{kovatchev-etal-2018-warp} annotation interface for the annotation. Following \citet{gold-etal-2019-annotating}, we created two separate binary annotation tasks - one for entailment and one for contradiction.  For entailment, we included each pair twice, changing the order of P and H to reflect the directional nature of the relation. If a sentence was annotated as not-entailment and not-contradiction, we marked it as ``neutral''. 

We use three annotators for each example. Following prior work \cite{marelli-etal-2014-sick,gold-etal-2019-annotating}, we calculated the agreement as the average \% of annotators that voted for the majority label. We obtained 86.9\% agreement for the ``entailment'' relation and 85.6\% agreement for the ``contradiction'' relation. We also calculated the Fleiss' kappa score, obtaining a ``moderate agreement'' of 55.\footnote{The lower  kappa is likely due to the label imbalance in the corpus: only about 22\% of the pairs contain entailment and 20\% contain contradiction.} Our agreement and label distribution of labels are consistent with the results reported by \citet{gold-etal-2019-annotating} for 10 annotators. Since 56\% of the pairs were labeled ``neutral'', we kept all ``entailment'' and ``contradiction'' pairs and randomly downsampled the ``neutral'' to obtain a balance between the classes. The \textit{annotate} portion contains 1,290 pairs.

\subsection{Generating Neutral Pairs}\label{i:c:neu}

Using our \textit{generation} strategy, we created ``entailment'' and ``contradiction'' pairs. Using our \textit{rewrite} strategy, we created pairs with all three relations. However the ``neutral'' class was underrepresented compared to the other two. To create a balanced dataset for training automated systems, we needed a separate strategy to introduce more pairs with the ``neutral'' label. In this subsection, we describe four different rule-based strategies that we used to generate ``neutral'' pairs in an automated manner.  

\paragraph{Shuffling existing P and H (same topic)} 
We matched each premise to two random hypotheses, generated for different premises on the same topic. 

\paragraph{Matching existing P with random contexts  (same topic)} In Section \ref{i:corp:ext}, we kept a ``context'' paragraph for each premise that we extracted. We matched each premise to two random ``contexts'' on the same topic. We then randomly selected a sentence from each of those contexts.


\paragraph{Matching existing H with random contexts (same topic)} Similar to the previous strategy, we randomly matched each hypothesis to a sentence from a ``context'' paragraph on the same topic.

\paragraph{Shuffling existing P and H (different topics)} 
Typically, the premise and the hypothesis have at least some degree of semantic similarity. We argue that an automated NLI solution should be able to label unrelated pairs. We created a small fixed number of unrelated pairs by matching texts and hypotheses from different topics. 

\paragraph{Validating neutral pairs} We selected 240 pairs, 60 from each of the four strategies, to manually validate the quality of the ``neutral'' pairs. One of our ``sentence generators'' performed a two-stage annotation on each pair. At the first stage they annotated \textit{``whether the premise and hypothesis are semantically releated''}. At the second stage they annotated \textit{``whether the meaning relation is neutral, despite a potential semantic relatedness''}. 55\% of the ``neutral'' pairs had some semantic relation (e.g., shared topic or named entities), and 26\% had a \textit{strong} relation. 
237 out of the 240 examples (98.75\%) were annotated as ``neutral''. Two pairs were found to be ``entailment'' and one - ``contradiction''. 

\paragraph{The ``neutral'' portion of the corpus} Through generation and downsampling, we obtain a total of 1,291 ``neutral'' sentences for \corp. We use 298 of them to re-balance the \textit{rewrite} portion of the corpus and the remaining 1,893 to complete the \textit{generation} portion of the corpus. 

\section{\corp}\label{inferes:cs}

We combined the examples from all corpus creation strategies to create \corp~- a corpus of NLI for Spanish containing 8,055 text pairs. 
Table \ref{i:tab:stratstat} shows the distribution of pairs and labels based on the creation strategy. Note that for \textit{generation} and \textit{rewrite} strategies, the ``neutral'' examples were at least in part generated automatically to ensure label balance. For ``annotate'' the ``neutral'' pairs are naturally occurring.
More than half of the corpus, 5,029 text pairs, was created using the \textit{generate} strategy. This is the core part of the corpus, in which we have incorporated multiple strategies for ensuring the quality and the linguistic diversity of the examples. 
1,716 pairs were created using the \textit{rewrite} strategy and 1,290 pairs were generated using the \textit{annotate} strategy.
All six topics are represented roughly equally. In the \textit{generate} and \textit{rewrite} portions, we aimed to ensure that each original premise has the same number of hypotheses, distributed equally across relations.

\begin{table}[h]
    \centering
    \begin{tabular}{|l|l|l|l|l|}
        \hline
        \textbf{Strategy} & \textbf{Pairs} & \textbf{Ent} & \textbf {Cnt} & \textbf{Neu} \\
        \hline
        Full        & 8,055     & 2,399 & 2,687 & 2,969 \\ 
        Generate    & 5,029*    & 1,574 & 1,582 & 1,893* \\
        Rewrite     & 1,716*    & 398   & 712   & 606* \\
        Annotate    & 1,290     & 427   & 393   & 470 \\
        \hline
    \end{tabular}
    \caption{Distribution of labels \corp by strategy}
    \label{i:tab:stratstat}
\end{table}

We measured the vocabulary size and the lexical overlap between the premise and hypothesis. The full \corp~has a vocabulary size of 12,877 unique types. On average, 22.6\% of the tokens from the premise also appear in the hypothesis. 33.4\% of the tokens from the hypothesis also appear in the premise. The two numbers differ since we count the number of non-unique tokens, including repetition, and we normalize them using a different denominator (length of premise/hypothesis). The overlap is comparable with prior work for English (20\% and 38\% for MNLI and 18\% and 34\% for linguist-in-the-loop NLI).

\section{Machine Learning Experiments}\label{inferes:exp}

To demonstrate the utility of \corp, we carried out a set of machine learning experiments. The design of \corp allows us to test NLI models under a variety of conditions: standard train/test split, hypothesis-only condition, performance on negation-based adversarial examples, and performance by topic in- and out-of-distribution.

\paragraph{Machine learning models} We used two transformer based models, pretrained for Spanish: the multilingual version of BERT \cite{devlin-etal-2019-bert} and the Spanish version of BERT, BETO \cite{CaneteCFP2020}. We used the version of the models available on HuggingFace \cite{wolf-etal-2020-transformers} as of May 2022 and finetuned them on \corp. After experimenting with different hyperparameter settings, we empirically found the best performance using a PolinomialDecay learning rate scheduler and training the model for five epochs. We kept the rest of the hyperparameters at their default values and used ADAM optimizer\footnote{The code for the experiments and all hyperparameters are available at \url{https://github.com/venelink/inferes}}. All reported results are the average of five different random initializations.

\begin{table}[h]
    \centering
    \begin{tabular}{|l|r|r|r|}
        \hline
        \textbf{Condition} & \textbf{MB} & \textbf{mBERT} & \textbf{BETO}\\ \hline
        Full Dataset    & 36.8 & 69.6 & 72.8 \\ 
        Hyp. Only       & 36.8 & 38.8 & 42.3 \\
        Adv. Negation   & 41.5 & 52 & 51.2 \\ \hline
    \end{tabular}
    \caption{Performance of multilingual-BERT and Spanish BERT (BETO) on \corp~ across different conditions. MB: ``majority baseline''. `Full'': standard train/test split.``Hyp. Only'': hypothesis-only. ``Adv. Negation'': negation-based adversarial examples. }
    \label{i:tab:full_results}
\end{table}

\paragraph{Full corpus performance} In a standard ``full corpus'' condition, we used 80\% of \corp for training, 10\% for validation, and 10\% for testing. We use the examples generated by all three strategies for both training and testing. As shown in Table \ref{i:tab:full_results}, BETO obtained 72.8\% accuracy and outperformed mBERT (69.6\%). While both models reached a fair performance on the test set, the results were much lower than the ``super-human'' performance that state-of-the-art transformer models obtain on popular benchmarks for English. For a reference, the official performance for the Spanish portion of XNLI is 82\% for BETO-cased, and 78.5\% for mBERT\footnote{See \url{https://github.com/dccuchile/beto}}.

\paragraph{Hypothesis-only performance} In the ``hypothesis-only'' condition, the models do not have access to the ``premise'' during training or testing. NLI explores the meaning relation between the two texts - the premise and the hypothesis. If a model is exposed only to one of the texts, its performance should not exceed that of a random guess, roughly equal to predicting the most common class.

Prior work has shown that existing datasets for English contain a large number of ``annotation artifacts'' and models are able to obtain much higher performance than chance. 
\citet{poliak-etal-2018-hypothesis} report 55\% accuracy for the ``hypothesis-only'' condition on the MNLI corpus using non-transformer models, an increase of 20\% over the majority baseline. \citet{parrish-etal-2021-putting-linguist} show that their ``linguist-in-the-loop'' approach is less biased in the hypothesis-only condition, but they also report accuracy over 50\% on the full dataset using a ROBERTA transformer. 

In the ``hypothesis-only'' condition on \corp, BETO obtained 42.3\%, and mBERT -- 38.8\%, which is respectively 5.5\% and 2\% higher than the majority baseline. The relatively small improvement over the majority baseline indicates that the hypothesis-only artifacts in \corp are substantially fewer than in previous work.

\paragraph{Performance on negation-based adversarial examples}  For this experiment, we trained the models on the \textit{generation} and \textit{annotation} portion of the corpus and evaluated them on \textit{rewrite}. The setup is similar to the one from \citet{hossain-etal-2020-analysis}. The performance of mBERT and BETO drops significantly when facing adversarial examples (See Table \ref{i:tab:full_results}). The models obtain 52\% and 51.1\% accuracy, about 10\% higher than the majority baseline. \citet{hossain-etal-2020-analysis} report that on two of the three datasets they use, BERT performs worse than the majority baseline. Our experimental setup is arguably more difficult, since \corp contains multiple negation strategies rather than just negating the main verb.
The \textit{rewrite} portion of the corpus can provide further insight into the use of negation in NLI and we hope it would facilitate further research and improvement in the area.

\paragraph{In- and Out-of-distribution generalization by topic} We also carried out a set of experiments to determine the ability of models to generalize across the six different topics. We evaluated the models in two different conditions.

\begin{table}[h]
    \centering
    \begin{tabular}{|l|r|r|r|r|}
        \hline
        \textbf{Top} & \multicolumn{2}{c|}{\textbf{mBERT}} & \multicolumn{2}{c|}{\textbf{BETO}} \\ \hline
        & IND & OOD & ID & OOD \\ \hline
        All        & \multicolumn{2}{c|}{69.6}   & \multicolumn{2}{c|}{72.8} \\
        $1$       &$73.1\pm4$&$67.9\pm2$&$78.5\pm4$&$73.8\pm.6$ \\
        $2$       &$68.1\pm4$&$69.9\pm1$&$74.6\pm3$&$72.0\pm.7$ \\
        $3$       &$70.8\pm3$&$70.9\pm1$&$74.2\pm3$&$73.3\pm.7$ \\
        $4$       &$71.6\pm2$&$68.9\pm.7$&$69.4\pm5$&$69.7\pm.7$ \\
        $5$       &$77.1\pm3$&$76.1\pm.5$&$78.3\pm4$&$77.2\pm.9$ \\
        $6$       &$65.3\pm4$&$69.8\pm.9$&$68.6\pm5$&$69.6\pm.9$ \\ 
	\hline
    \end{tabular}
    \caption{In-distribution (ID) and Out-of-distribution (OOD) performance of mBERT and BETO on different topics within \corp. ID: model trained on data covering all 6 topics. OOD: model trained on 5 topics and evaluated on the unseen 6th. 1 (Picasso), 2 (Columbus), 3 (Olympics), 4 (Videogames), 5 (EU), 6 (USSR).}    \label{i:tab:crosst}
\end{table}

The in-distribution (ID) condition is an extension of the ``full corpus'' condition. We split the full corpus containing all six topics in an 80/10/10 ratio. Then, when evaluating the performance of the models, we split the test set in six sub-sets, based on their topic and we measured the model accuracy on each sub-set. To ensure that the variation of the model performance is not due to a sampling bias, we re-trained each model five times, using a different 80/10/10 random split each time. In Table  \ref{i:tab:crosst}, we report the average accuracy across the five different splits. 
Both models obtained the highest ID performance on the topics of ``Picasso'' and ``The European Union'' and the lowest ID performance on ``The USSR''. 

For the out-of-distribution (OOD) condition, we designed a transfer learning experiment, in which we trained mBERT and BETO on five of the topics and evaluate on the sixth. The results presented in Table \ref{i:tab:crosst} demonstrate that the models are able to generalize well across the topics, even in a transfer learning setup. For both models, the OOD performance on most topics drops between 1\% and 5\% compared to ID. The performance on ``Olympics'' for mBERT and on ``videogames'' for BETO was almost identical between conditions. For ``The USSR'', both models obtained higher performance for OOD. We inspected the matter further and noticed that due to the corpus size, the ID random split is not very stable (the ID test set only contains between 100 and 120 instances of each topic) and the average can be affected by outliers. The OOD results are much more stable due to the test set having over 1,000 examples per topic. This can be seen in the difference in standard deviation: between 2\% and 5\% for ID, and below 1\% for OOD. Our experiments demonstrate that the models are able to generalize well even to topics they have never seen during training, a promising finding for the overall generalizability of NLI models.


\section{Discussion}\label{inferes:diss}

In Section \ref{inferes:corp}, we formulated three main objectives behind \corp. In this section, we want to revisit  those objectives and briefly discuss the importance of our work for the NLP and NLI communities.

Our \textbf{first objective} was \textit{``To create a native NLI dataset for the Spanish language''}. To the best of our knowledge, \corp is the first native NLI dataset for Spanish which is not adapted or translated. We described the creation process and validated that it can be used to train different machine learning models. We have successfully contributed a new resource to the Spanish NLP community and we hope that \corp can facilitate the further creation of tools and resources for that language.

Our \textbf{second objective} was \textit{``To promote better data quality and corpus creation practices.''}. We proposed, implemented, analyzed, and compared several different strategies for creating text pairs. 
The resulting dataset proves non-trivial to state-of-the-art NLP models with an overall accuracy in the low 70s. This leaves a lot of room for improvement and future research. The results using a ``hypothesis-only'' baseline indicate that \corp contains fewer annotation artifacts than prior work. At the same time, models trained on the dataset are able to generalize well across different topics, even in our ``out-of-distribution`` condition. Overall, we can conclude that \corp is of high quality and achieves the objective of promoting better data by design.

Our \textbf{third objective} was \textit{``To facilitate the research on negation and coreference in the context of NLI.''}. Our \textit{rewrite} strategy was focused on creating naturally occurring contrastive and adversarial examples based on negation and coreference. We followed prior work on evaluating systems' performance and demonstrated that those examples are non-trivial to solve. However, the two models are still able to outperform the majority baseline by over 10\%. These findings indicate that the problem is not unsolvable and the models are learning  something about complex negation from the data. \corp can facilitate the resaerch of negation in Spanish both in the context of NLI and in isolation. We leave quantifying the importance of coreference in the \textit{rewrite} section for future work.

Overall, we have achieved all our objectives: 1) we created a novel dataset for Spanish; 2) we used different generation and annotation strategies to obtain a challenging corpus with fewerr annotation artifacts; 3) we created a set of high-quality contrastive and adversarial examples based on negation and coreference. We believe that \corp is an important contribution to Spanish NLP, and also to researchers interested in NLI, negation, and coreference. We hope that this dataset can be used to train and evaluate more accurate automated systems, but also to better understand the nature of those linguistic phenomena.

\section{Conclusions}\label{inferes:conc}

We presented \corp - a new corpus of Natural Language Inference for Spanish. To the best of our knowledge, this is the first original Spanish NLI corpus that is not a translation or an adaptation of an existing dataset. We explored several different strategies for corpus creation and put the emphasis on creating diverse and non-trivial examples, that are also linguistically interesting. More specifically, we created contrastive and adversarial examples involving complex negation and coreference.

We provided two baseline transformer-based systems finetuned on the dataset. We demonstrated that \corp is challenging and contains fewer anotation artifacts than prior work. We also evaluated the performance of automated systems on adversarial examples and the ability of the models to generalize across topics in- and out-of-distribution. The results validated the quality and the difficulty of the corpus. \corp leaves a room for analysis and improvement.

Our work opens several directions for future work: studying and improving the performance of NLI models for Spanish; expanding the research on negation in Spanish, and specifically the complex and lexical negation; evaluating the importance of coreference in the context of NLI. We believe that \corp will be useful both to Spanish researchers and to the general NLP community. 

\section*{Acknowledgements}

We want to thank M. Ant\'{o}nia Mart\'{i} and Montse Nofre for their suggestions and support for this work. We are grateful to Patricia Grau Francitorra, Eugenia Verjovodkina, V\'{i}ctor Bargiela Zotes, and Xavier Bonet Casals for the annotations and the discussions about the annotation process. We also want to thank the anonymous reviewers for their feedback and suggestions.

This work was partially funded by the project ``FairTransNLP-Language: Analysing Toxicity and Stereotypes in Language for Unbiased, Fair and Transparent Systems (PID2021-124361OB-C33)'' funded by Ministerio de Ciencia e Innovación (Spain) and co-funded by the European Regional Development Fund (FEDER). Most of this work was carried out during the first author's APIF grant at the University of Barcelona.

\bibliography{anthology,custom}
\bibliographystyle{acl_natbib}

\end{document}